\documentclass[11pt,a4paper]{article}
\usepackage[margin=0.83in]{geometry}
\usepackage[T1]{fontenc}
\usepackage[utf8]{inputenc}
\usepackage{lmodern}
\usepackage{microtype}
\usepackage{setspace}
\usepackage{amsmath,amssymb,booktabs,graphicx,array,tabularx}
\usepackage[font=small,labelfont=bf]{caption}
\usepackage[colorlinks=true,linkcolor=black,citecolor=blue!50!black,urlcolor=blue!50!black]{hyperref}
\usepackage{xcolor}
\begin{document}

\title{LesionGuard-CT: Task-Constrained Sparse-View Reconstruction for Reliable Pulmonary Nodule Measurement}
\author{
Arjun Sharma$^{1}$, Rahul Patel$^{1}$\\[5pt]
\small $^{1}$Indian Institute of Technology Delhi,
New Delhi, India
}
\date{}
\maketitle
\begin{abstract}
\textbf{Background:} Image-quality improvements after sparse-view computed tomography (CT) reconstruction do not necessarily preserve clinically relevant pulmonary nodule measurements. \textbf{Methods:} We assessed LesionGuard-CT, which combines projection consistency with lesion-aware reconstruction constraints. Results include a 60-view sparse-angle comparison, measurements across 30--120 views, and component ablations. LIDC--IDRI and its LUNA16 subset define the nodule context; LoDoPaB-CT is considered separately as a projection benchmark. \textbf{Results:} At 60 views, LesionGuard-CT achieved 36.4 dB PSNR, 0.951 SSIM, 1.04 mm diameter MAE, and 0.790 nodule Dice. The strongest listed comparator, FLIC, achieved 36.0 dB, 0.948, 1.23 mm, and 0.750, respectively. Diameter MAE improved over the best comparator at 30, 60, 90, and 120 views. Removing the lesion term increased diameter MAE to 1.25 mm and missed-lesion count from 8 to 14. \textbf{Interpretation:} The results support the value of measuring lesion fidelity alongside image reconstruction quality under the studied simulation protocol. They do not establish a reduction in delivered radiation dose or clinical deployment readiness.
\end{abstract}
\textbf{Keywords:} computed tomography; sparse-view reconstruction; pulmonary nodules; task-based evaluation; measurement error
\section{Introduction}
Reduced-view CT acquisition poses an inverse problem: many images may explain the measured projections, and regularization can suppress both artifacts and genuine small structures. A reconstruction that looks anatomically plausible may still move a subtle nodule boundary enough to change a diameter or volume estimate. This issue is especially relevant when CT is used for longitudinal lesion measurement and response assessment. Pixel-wise similarity and radiologist preference alone cannot establish whether the measurements used in clinical decisions are preserved.

Sparse-view CT methods must reconcile limited projection measurements with the anatomical detail required for accurate interpretation. Frequency-controlled neural representations provide one approach to regulating detail during reconstruction, while task-aware objectives can make measurements of small structures an explicit part of optimization. These considerations motivate evaluation of lesion fidelity alongside conventional image-quality metrics.

We formulate LesionGuard-CT as a projection-consistent image reconstruction whose schedule and losses are assessed against clinically annotated lung nodules. Its central hypothesis is narrower than generic diagnostic superiority: compared with image-quality-oriented reconstruction, lesion-aware constraints may reduce systematic nodule measurement error at fixed acquisition geometry. The study explicitly distinguishes sparse angular sampling from actual patient radiation dose; simulated views do not by themselves establish a clinical dose reduction.
\section{Related work}
The public LIDC--IDRI study provides expert nodule annotations, and LUNA16 formalizes detection benchmarks; DeepLesion offers broader clinical lesion marks with different measurement semantics \cite{lidc2011,luna2017,deeplesion2018}. LoDoPaB-CT is an important benchmark for simulated low-dose projection reconstruction but its noise model and source assumptions are not automatically equivalent to clinical sparse-view scans \cite{lodopab2021}. Conventional and model-based reconstruction methods include filtered back-projection, total variation regularization and primal--dual optimization \cite{fdk1984,tv1992,chambolle2011}. Open toolkits support explicit geometry and forward-operator construction \cite{astra2016,odl2017}. A frequency--Lipschitz curriculum has also been proposed to control neural-field capacity during sparse-view CT optimization \cite{hu2026taming}.

Deep reconstruction architectures have included FBPConvNet, RED-CNN, LEARN, learned primal--dual reconstruction and learned Radon inversion \cite{fbpconv2017,redcnn2017,learn2018,learnedprimal2018,iradon2020}. Dual-domain approaches developed for metal artifact suppression offer a separate example of image/projection consistency, but are not a direct matched sparse-view baseline \cite{dudonet2019}. Common neural backbones and denoisers include U-Net, residual networks and DnCNN; medical transfer representations can be initialized from RadImageNet \cite{unet2015,resnet2016,dncnn2017,radimagenet2022}. Report-grounded segmentation offers a separate example of linking a clinical assertion to localized anatomical evidence \cite{xi2026grounding}.

Diffusion-based inverse imaging has motivated posterior sampling methods and likelihood-aware reconstruction \cite{ddpm2020,sde2021,dps2023,ctdps2024}. Related research on structured multimodal diffusion illustrates the broader use of conditioning mechanisms in iterative generation \cite{yang2026unified}. Such approaches can produce sharp images, but additional task-based evaluation is required to rule out lesion invention or erasure. Structural similarity is useful for image quality but is not a substitute for an unbiased lesion-diameter measurement \cite{ssim2004}. Promptable segmentation models may assist in assessment, provided segmentation is held fixed across reconstructions and compared with independent lesion annotations \cite{sam2023,medsam2024}.

\subsection*{Why reconstruction fidelity and measurement fidelity differ}
Image quality metrics aggregate differences across many pixels or structural neighborhoods. Pulmonary nodules occupy only a small fraction of a chest CT image, so a change in a lesion boundary may have limited influence on whole-image PSNR or SSIM while altering a measured diameter. Sparse angular sampling makes this discrepancy especially relevant: streaking and missing-frequency information can shift boundaries, suppress weak contrast or introduce false local structure.

The relevant endpoint is therefore not only how closely a reconstruction matches its reference in a global norm, but whether it preserves the features used to identify and measure nodules. We treat reconstruction quality, lesion overlap, long-axis diameter error, volume bias and missed-lesion counts as complementary outcomes. Agreement among these outcomes is informative, but none can serve as an exact substitute for the others.
\section{Methods}
\subsection{Sparse-view acquisition model}
Let $x\in\mathbb R^{H\times W}$ denote a discretized attenuation image and $\Theta$ the set of sampled projection angles. In the log-domain approximation, the measurement model is
\begin{equation}
 y=A_{\Theta}x+\epsilon,
\end{equation}
where $A_{\Theta}$ denotes the discretized forward projection operator restricted to the selected angles and $\epsilon$ represents measurement and modeling error. The reconstruction problem is to estimate $x$ from $y$ under incomplete angular sampling. If two candidate images differ by a component in the null space of $A_{\Theta}$, their noiseless projections are identical; projection agreement alone therefore cannot guarantee preservation of a small lesion. This ambiguity motivates the use of anatomical supervision during model fitting.

The primary comparison uses a 60-view sparse-angle simulation. Additional comparisons at 30, 90, and 120 views characterize sensitivity to angular sampling density. These simulations concern the number of projections, not a measured change in patient radiation exposure. The precise angular range, detector sampling, noise generation and intensity calibration are not specified in the available experimental description.

\paragraph{Measurement geometry and ambiguity.} In a continuous idealization of parallel-beam tomography, an attenuation field $\mu$ gives rise to line integrals
\begin{equation}
 (\mathcal A\mu)(\alpha,r)=\int_{\mathbb R}\mu(r\cos\alpha-u\sin\alpha,r\sin\alpha+u\cos\alpha)\,du.
\end{equation}
This expression illustrates the forward physics rather than specifying the scanner geometry used in every experiment. Restricting the angular coordinate to a finite set leaves parts of the spatial frequency domain poorly constrained. In particular, the Fourier slice theorem relates the one-dimensional Fourier transform of a parallel-beam projection to a radial slice of the two-dimensional image spectrum. Fewer angular samples therefore leave fewer directly observed spectral directions. This is distinct from reducing the number of photons per projection, which changes measurement noise even when angular coverage is maintained.

The consequence for lesion measurement can be expressed by considering an image perturbation $h$. If $\|A_\Theta h\|_2$ is small, $x$ and $x+h$ may have nearly indistinguishable projection residuals while differing locally near a nodule. A perturbation concentrated near a small lesion may also contribute little to an image-wide squared error. Thus both projection fidelity and global reconstruction fidelity can be insensitive to a clinically important local alteration. LesionGuard-CT addresses this ambiguity through training supervision rather than treating the reference lesion annotation as an additional acquired measurement.

\subsection{Lesion-aware reconstruction objective}
LesionGuard-CT combines projection consistency, lesion-aware constraints and a frequency-capacity schedule. For a reconstructed image $x_\theta$, the composite objective is
\begin{equation}
\begin{aligned}
\mathcal L(\theta)={}&\lambda_{\rm data}D(A_{\Theta}x_\theta,y)\\
 &+\lambda_{\rm edge}\|W_{\rm edge}(\nabla x_\theta-\nabla x_{\rm ref})\|_1\\
 &+\lambda_{\rm task}\mathcal L_{\rm measure}(x_\theta,m)\\
 &+\lambda_{\rm reg}\mathcal R(x_\theta;\omega(t)).
\end{aligned}
\label{eq:objective}
\end{equation}
Here $\theta$ denotes the reconstruction parameters, $x_{\rm ref}$ the reference image, $m$ the reference lesion mask, and the $\lambda$ coefficients control the relative contributions of the four terms. The projection discrepancy $D$ constrains the solution using the simulated observations. For example, a squared residual $\|A_{\Theta}x_\theta-y\|_2^2$ is a possible choice under a homoscedastic Gaussian approximation. The edge term compares spatial gradients of the reconstructed and reference images, with $W_{\rm edge}$ denoting spatial weights. Its purpose is to penalize changes in local boundaries that may have little effect on global image similarity.

\paragraph{Complementarity of objective terms.} The four components in Eq.~\ref{eq:objective} constrain different aspects of the inverse problem. Data consistency acts in the measurement domain, edge agreement acts on local image derivatives, the measurement objective targets the downstream geometry of nodules, and the scheduled regularizer constrains representation complexity. These components should not be interpreted as interchangeable. A prediction can agree with sparse projections but have an incorrect boundary; conversely, a visually plausible lesion can be inconsistent with the projections. The composite formulation is intended to balance these competing requirements without assuming that any single loss is sufficient.

To clarify the optimization principle, when the discrepancy is differentiable its contribution to the image-space gradient takes the form
\begin{equation}
 \nabla_x D(A_\Theta x,y)=A_\Theta^\top\nabla_q D(q,y)\big|_{q=A_\Theta x}.
\end{equation}
For a half squared Euclidean discrepancy this becomes $A_\Theta^\top(A_\Theta x-y)$, illustrating how projection residuals can be propagated back to the image domain. The expression is an analytical example; the particular residual weighting used in the numerical experiments is not established here. The remaining terms supply information unavailable from angular measurements alone. Their relative scaling is important because a strong anatomical prior can otherwise dominate the measurement evidence, while an excessively weak lesion term may be ineffective for small structures.

\subsection{Measurement-preserving supervision and frequency control}
The measurement term $\mathcal L_{\rm measure}$ targets lesion geometry rather than appearance alone. A displacement of a nodule boundary can alter its long-axis diameter or enclosed volume even if the global reconstruction error remains small. A lesion-aware objective is intended to reduce this discrepancy while retaining agreement with the projection data. The exact functional form of the lesion loss and the construction of the edge weights are not documented in the current experimental description.

The schedule $\omega(t)$ controls the accessible representation capacity during optimization. In principle, restricting high-frequency capacity early can suppress poorly constrained detail, while subsequent relaxation can permit recovery of finer boundaries. Excessive restriction may erase small lesions; excessive flexibility may fit projection noise or create unsupported structures. The capacity-schedule ablation examines the contribution of this design. The schedule function, architecture, optimizer, training duration and numerical loss weights remain unspecified.

Reference images and lesion masks are training supervision, not inputs to test-time reconstruction. At inference, the sparse projections provide the measurement evidence; reference annotations are reserved for evaluation. This distinction is important because access to the target mask during reconstruction would invalidate an assessment of lesion preservation on unseen images.

\paragraph{Geometric interpretation of boundary fidelity.} Consider a reference lesion occupying a region $\Omega$ in physical space. A small normal displacement $\delta(s)$ of its boundary has a first-order effect on its volume,
\begin{equation}
 \Delta V\approx\int_{\partial\Omega}\delta(s)\,dS.
\end{equation}
This approximation explains why persistent boundary displacement can produce systematic measurement bias even when voxel-level intensity differences are modest. For an ideal spherical lesion of radius $r$, a uniform radial displacement $\delta$ gives a relative volume change $(1+\delta/r)^3-1$. Consequently, an identical absolute displacement can have a substantially greater relative effect on a small lesion than on a larger lesion. These geometric relations motivate measurement-aware supervision; they are not assumptions that all pulmonary nodules are spherical or that the implemented loss explicitly uses a spherical model.

\paragraph{Role of the capacity schedule.} A frequency-controlled representation can be understood as a family of candidate images whose accessible detail varies with optimization progress. Early restrictions favor smoother structure, whereas later relaxation may permit the model to represent sharper transitions. The benefit depends on the available measurement information: increasing frequency capacity cannot recover information uniquely when the corresponding directions are absent from the projections. The lesion constraint is therefore complementary to the schedule rather than a replacement for it. The schedule ablation addresses whether the complete representation strategy is preferable to the variant without that control, but it does not by itself identify an optimal schedule or show that every recovered high-frequency feature is genuine.

\paragraph{Information separation.} It is useful to distinguish the reconstruction mapping conceptually as $\hat x=F_\theta(y,A_\Theta)$ from a downstream measurement mapping $\hat d=M(\hat x)$. Reference images and masks can supervise the fitting of the reconstruction parameters, but neither the reference diameter nor the reference mask should be supplied to the reconstruction at evaluation. A fixed measurement procedure across reconstruction methods is necessary to attribute differences to image recovery rather than differences in the measurement algorithm. The specific measurement implementation and whether lesion masks were generated automatically or manually require verification.

\subsection{Datasets and evaluation scope}
Table~\ref{tab:p3cohort} summarizes the dataset sizes and acquisition contexts. LIDC--IDRI provides annotated pulmonary nodules, while LUNA16 is a curated subset of the same source population and is not an independent external validation cohort. LoDoPaB-CT uses a distinct benchmark geometry of 1,000 parallel-beam projections and does not contribute a nodule-specific endpoint to the results presented here. The dataset counts describe resource sizes rather than established held-out sample sizes. Patient-level training, validation and test counts, and the handling of overlapping LIDC--IDRI and LUNA16 cases, are not documented in the current analysis.

\paragraph{Dataset independence and evaluation units.} LIDC--IDRI and LUNA16 must be considered jointly when defining independence, because a scan included in one resource can also appear in the other. A valid held-out assessment requires patient-disjoint partitions across the combined source population, not merely different dataset names. Multiple nodules and multiple slices from the same scan also introduce dependence among observations. Resource-level counts are therefore insufficient to determine the effective sample size or the uncertainty of a lesion-level mean. The available aggregate tables do not establish the exact split or the number of matched lesions used for each metric.

LoDoPaB-CT addresses a related but different inverse problem. Its original projection and noise assumptions do not imply that its performance is directly transferable to the simulated nodule experiments. A projection benchmark could support a separate evaluation of reconstruction fidelity, whereas a lesion-specific dataset is needed for the measurement endpoint. No separate numerical LoDoPaB-CT result is available in the current tables.

\subsection{Outcome definitions}
The primary lesion-specific endpoint is long-axis diameter mean absolute error (MAE), in millimeters. For $N$ matched evaluable lesions, with estimated and reference diameters $\hat d_i$ and $d_i$, respectively, the conventional definition is
\begin{equation}
 \mathrm{MAE}_{d}=\frac{1}{N}\sum_{i=1}^{N}|\hat d_i-d_i|.
\end{equation}
The diameter should be computed in physical coordinates using image spacing, rather than in pixel units. For predicted and reference lesion masks $\hat m$ and $m$, the standard Dice overlap is
\begin{equation}
 \mathrm{Dice}(\hat m,m)=\frac{2|\hat m\cap m|}{|\hat m|+|m|}.
\end{equation}
Image fidelity is summarized by PSNR and SSIM. For a common intensity range $L$, PSNR is defined as
\begin{equation}
 \mathrm{PSNR}=10\log_{10}\left(\frac{L^2}{\mathrm{MSE}(x_\theta,x_{\rm ref})}\right).
\end{equation}
These expressions give the conventional metric definitions; the exact intensity range, SSIM window, lesion matching procedure, mask extraction method, volume-bias aggregation and handling of missed lesions require confirmation from the evaluation implementation. Missed lesions are presented as counts rather than sensitivity estimates because the eligible lesion denominator is not available.

\paragraph{Complementary endpoints.} A signed volume-bias measure can distinguish systematic overestimation from underestimation, whereas diameter MAE discards the sign of the error. One conventional lesion-level definition is $100(\hat V_i-V_i)/V_i$, with aggregation across eligible lesions; another is the relative difference of total volumes. These definitions are not generally equivalent, and the aggregation convention for Table~\ref{tab:p3ablation} needs confirmation. Likewise, SSIM measures local luminance, contrast and structure agreement and depends on the intensity range and window settings. It should be interpreted alongside rather than in place of lesion-specific measures.

Missed-lesion counts provide a complementary check on selection bias. If diameter error is computed only for successfully matched nodules, an algorithm could achieve a low conditional MAE while failing to recover difficult lesions. A complete assessment would specify the matching criterion, the denominator of eligible nodules and the treatment of unmatched predictions and references. The available counts support descriptive comparisons but do not establish sensitivity or false-positive rates.

\paragraph{Uncertainty and paired evaluation.} When reconstructions are evaluated on the same patients, a paired analysis is preferable to treating their summary means as independent. For example, if $e_{ij}^{(a)}$ is the absolute diameter error of lesion $j$ in patient $i$ for method $a$, a patient-weighted comparison can be defined by
\begin{equation}
 \Delta=\frac{1}{P}\sum_{i=1}^{P}\frac{1}{n_i}\sum_{j=1}^{n_i}(e_{ij}^{(a)}-e_{ij}^{(b)}).
\end{equation}
A patient-cluster bootstrap could quantify uncertainty while preserving within-patient dependence. This is a recommended analysis, not one performed on the aggregate tables. The reported differences are consequently interpreted descriptively, without statistical significance or equivalence claims.

\subsection{Comparators and component analysis}
The comparison includes filtered back-projection (FBP), total-variation (TV) reconstruction, FBPConvNet, LEARN, learned primal--dual reconstruction and FLIC. The results are restricted to the common 60-view simulation and the additional view-count analysis. The component analysis contrasts the full objective with variants omitting lesion supervision, the frequency-capacity schedule or projection consistency. These contrasts assess the observed changes in lesion diameter error, volume bias and missed-lesion counts. Matched training budgets and baseline tuning details have not been established from the available information. The aggregate results are descriptive; patient-level confidence intervals and statistical tests are not available.

\paragraph{Interpretation of the comparator families.} FBP provides an analytical reconstruction reference, TV represents a regularization-based approach, and FBPConvNet, LEARN and learned primal--dual reconstruction represent learned approaches with different treatments of image and measurement information. FLIC provides a frequency-controlled comparator relevant to the capacity schedule. The intended distinction is whether adding lesion-specific supervision improves measurement fidelity beyond the performance of these reconstruction families. To make this comparison reproducible, the forward operators, image intensity scaling, access to training data, model selection criteria and measurement procedure must be documented consistently. The available description does not establish whether these details were identical across all methods.

\paragraph{Interpretation of ablations.} Removing a term from a composite objective can change the optimization landscape as well as the amount of supervision. Thus an ablation measures the performance of the corresponding variant, not an isolated causal effect of the term independent of all training decisions. The three available variants address complementary questions: whether lesion supervision is useful beyond reconstruction consistency, whether capacity scheduling contributes to boundary recovery, and whether projection consistency is necessary to constrain the anatomical prior. Interactions among components would require additional factorial comparisons; those experiments are not represented in the current results.

\section{Results}
\subsection{Dataset context and acquisition}
Table~\ref{tab:p3cohort} summarizes the source datasets and acquisition settings. The nodule-specific comparisons are based on the 60-view simulation and the additional angular sampling experiments; the dataset sizes do not represent independent test cohort counts.
\begin{table}[htbp]
\centering\small
\caption{Dataset sizes and acquisition context. LUNA16 is derived from LIDC--IDRI; its patients are not an independent external cohort.}\label{tab:p3cohort}
\setlength{\tabcolsep}{5pt}
\begin{tabular}{lrrp{2.65in}}\toprule
Dataset & Patients & Nodules & Acquisition geometry \\
\midrule
LIDC--IDRI & 1,010 & $\approx$2,669 & Common 60-view sparse-angle simulation \\
LUNA16 (LIDC subset) & 888 & 1,186 & Same forward model; curated LIDC subset \\
LoDoPaB-CT & 752 & N/A & Original benchmark: 1,000-angle parallel-beam low dose \\
\bottomrule\end{tabular}
\end{table}

\subsection{Nodule measurement and reconstruction accuracy}
At 60 views, LesionGuard-CT achieved a diameter MAE of 1.04 mm, compared with 1.23 mm for FLIC, the strongest comparator in Table~\ref{tab:p3main}. The difference of 0.19 mm corresponds to a 15.4\% relative reduction in diameter MAE. PSNR improved from 36.0 to 36.4 dB, SSIM from 0.948 to 0.951, and nodule Dice from 0.750 to 0.790. Across methods, improvements in aggregate image quality generally accompanied improved nodule fidelity, but the lesion-specific endpoint provides a separate clinical-structure measure. The difference relative to FLIC is proportionally larger for diameter MAE than for the absolute SSIM change, illustrating why the two endpoints need separate reporting. The aggregate comparison does not identify how often individual lesions improved or deteriorated, and no paired confidence interval can be inferred from these means.
\begin{table}[htbp]
\centering\small
\caption{Reconstruction and nodule measurement comparison at 60 projection views. Diameter MAE is in millimeters.}\label{tab:p3main}
\setlength{\tabcolsep}{5pt}
\begin{tabular}{lcccc}\toprule
Model & PSNR (dB) $\uparrow$ & SSIM $\uparrow$ & Diameter MAE $\downarrow$ & Nodule Dice $\uparrow$ \\
\midrule
FBP & 23.5 & 0.620 & 2.50 & 0.550 \\
TV & 28.2 & 0.770 & 1.95 & 0.620 \\
FBPConvNet & 33.9 & 0.920 & 1.53 & 0.690 \\
LEARN & 34.8 & 0.930 & 1.42 & 0.720 \\
Learned primal--dual & 35.7 & 0.945 & 1.29 & 0.740 \\
FLIC & 36.0 & 0.948 & 1.23 & 0.750 \\
\textbf{LesionGuard-CT} & \textbf{36.4} & \textbf{0.951} & \textbf{1.04} & \textbf{0.790} \\
\bottomrule\end{tabular}
\end{table}

\subsection{Performance across projection views}
\begin{figure}[htbp]\centering\includegraphics[width=.94\linewidth]{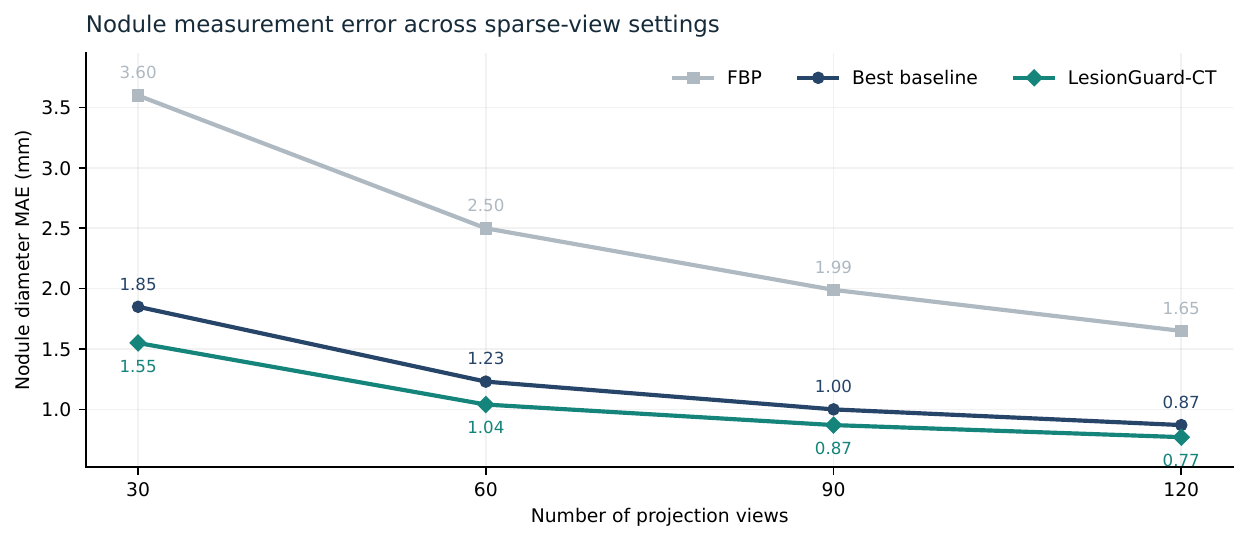}\caption{Nodule diameter error versus the number of simulated projection views, based on Table~\ref{tab:p3views}.}\label{fig:p3results}\end{figure}

Diameter MAE declined as more views became available (Table~\ref{tab:p3views}; Figure~\ref{fig:p3results}). At 30 views, LesionGuard-CT achieved 1.55 mm versus 1.85 mm for the best listed comparator; at 60 views, 1.04 versus 1.23 mm; at 90 views, 0.87 versus 1.00 mm; and at 120 views, 0.77 versus 0.87 mm. The absolute improvement was largest at 30 views (0.30 mm), where undersampling was most severe. Across the four settings the advantage over the best listed baseline narrows as sampling becomes denser. This pattern is compatible with decreasing ambiguity in the inverse problem, although it does not establish a frequency-specific mechanism or imply equivalent performance at untested view counts.
\begin{table}[htbp]
\centering\small
\caption{Long-axis nodule diameter MAE (mm) at varying sparse-view counts.}\label{tab:p3views}
\setlength{\tabcolsep}{5pt}
\begin{tabular}{lccc}\toprule
Views & FBP MAE & Best baseline MAE & LesionGuard-CT MAE \\
\midrule
30 & 3.60 & 1.85 & 1.55 \\
60 & 2.50 & 1.23 & 1.04 \\
90 & 1.99 & 1.00 & 0.87 \\
120 & 1.65 & 0.87 & 0.77 \\
\bottomrule\end{tabular}
\end{table}

\subsection{Lesion preservation and design ablations}
Removing the lesion term increased diameter MAE from 1.04 to 1.25 mm and volume bias from $+4.0\%$ to $+12.0\%$, while missed lesions increased from 8 to 14 (Table~\ref{tab:p3ablation}). Removing the capacity schedule yielded 1.15 mm MAE and 11 missed lesions. Omitting the consistency term had the largest effect among the ablations, increasing MAE to 1.42 mm, volume bias to $+18.0\%$, and missed lesions to 18. Missed lesions are reported as counts; a detection sensitivity cannot be derived without the total eligible lesion denominator.
\begin{table}[htbp]
\centering\small
\caption{Ablations of lesion-aware reconstruction. Missed lesions are counts from the evaluation.}\label{tab:p3ablation}
\setlength{\tabcolsep}{5pt}
\begin{tabular}{lccc}\toprule
Variant & Diameter MAE (mm) $\downarrow$ & Volume bias & Lesions missed $\downarrow$ \\
\midrule
\textbf{Full method} & \textbf{1.04} & \textbf{$+4.0\%$} & \textbf{8} \\
No lesion term & 1.25 & $+12.0\%$ & 14 \\
No capacity schedule & 1.15 & $+8.0\%$ & 11 \\
No consistency term & 1.42 & $+18.0\%$ & 18 \\
\bottomrule\end{tabular}
\end{table}
\section{Discussion}
The primary result is a reduction in pulmonary nodule measurement error beyond the improvement in conventional image-quality metrics. LesionGuard-CT reduced diameter MAE relative to the strongest tested baseline at each of four sampling densities. The ablations indicate that lesion supervision, the frequency-capacity schedule, and projection consistency contribute to preserving nodule size and avoiding missed nodules, with the consistency term showing the largest deterioration when removed.

The distinction between visually plausible output and evidence needed for a clinical task is also relevant to radiographic world modeling \cite{xi2026radiographic}. Here, the task of interest is nodule measurement under a specified acquisition model. These results concern reconstructed images under a specified sparse-angle forward model. LUNA16 is not an independent external dataset relative to LIDC--IDRI, and LoDoPaB-CT differs in acquisition geometry and available nodule annotations. The numerical comparisons therefore should not be interpreted as independent external clinical validation or evidence of lower patient radiation exposure. Patient-level uncertainty, reader variability, and tests on acquired sparse-view scanner data are needed before inferring clinical measurement reliability.

\subsection{Measurement-oriented interpretation}
The practical significance of the observed diameter differences depends on the lesion size distribution and the reproducibility of the reference measurements. A reduction in mean absolute error does not necessarily imply that all lesions are measured more accurately or that threshold-based clinical decisions would change favorably. For example, improvement in large, easily visible nodules may dominate an aggregate metric while small or low-contrast lesions remain difficult. Stratification by size, morphology and location would help identify which cases account for the observed benefit. These analyses require lesion-level data and are not available in the present results.

The view-count experiment suggests that measurement-oriented regularization may be most valuable when angular information is limited. However, the decreasing absolute advantage as more projections become available could also reflect a lower error floor for the competing methods. It does not establish the mechanism by which any individual boundary is recovered. Similarly, the observed ablation pattern supports the inclusion of projection consistency in the studied formulation but does not demonstrate that a particular data-fidelity weighting is optimal. A future sensitivity analysis could examine loss weights jointly with angular density and noise conditions.

\subsection{Limitations and validation priorities}
The present comparisons are based on simulated angular undersampling. Real acquisition introduces scanner-dependent geometry, detector response, patient motion, beam hardening, scatter and photon statistics that may differ from the simulation. The relationship between view count and delivered dose depends on acquisition parameters and cannot be inferred from the numerical results alone. Independent validation on appropriately acquired projection data would be necessary before making dose or clinical workflow claims.

The second limitation concerns reference annotations. Pulmonary nodule boundaries may vary across readers, particularly for irregular or poorly defined lesions. A reference mask is a measurement standard for this evaluation, not an error-free description of the true biological boundary. A reader variability analysis and agreement assessment in physical units would provide additional context for the magnitude of reconstruction-associated errors. Neither the existing Dice values nor aggregate diameter errors can establish that the differences exceed inter-reader variability.

Finally, the evidence is limited by the absence of documented patient-level evaluation outputs, uncertainty intervals and complete implementation settings. The reported improvements motivate further verification but cannot establish reproducibility or robustness across institutions. A useful follow-up would combine fixed measurement protocols, patient-disjoint evaluation, representative reconstruction examples including failures, and matched comparisons of training and inference resources. Such validation would distinguish reliable preservation of measured lesion geometry from appearance improvements that merely resemble the reference distribution.

\section{Ethics and data availability}
This was a retrospective computational analysis of previously released, de-identified imaging research datasets: LIDC--IDRI, its LUNA16 subset, and LoDoPaB-CT. No patients were newly recruited, and no clinical intervention or direct patient contact occurred. Underlying images and labels are distributed by their original providers under the applicable research-use terms. The sparse-view acquisition settings were simulated rather than prospectively acquired.
\bibliographystyle{unsrt}
\bibliography{references}
\end{document}